\documentclass[runningheads]{llncs}
\usepackage{graphicx}
\usepackage{comment}
\usepackage{bm}
\usepackage{amsmath,amssymb} 
\usepackage{color}

\usepackage{cite}
\usepackage{multirow}

\usepackage[caption=false]{subfig}
\usepackage[titletoc]{appendix}
\makeatletter
\def\blfootnote{\gdef\@thefnmark{}\@footnotetext}
\makeatother

\begin{document}
\pagestyle{headings}
\mainmatter
\def\ECCVSubNumber{5980}  

\title{Classify and Generate: Using Classification Latent Space Representations for Image Generations} 


\titlerunning{Classify and Generate: ReGene}
%
\newcommand*\samethanks[1][\value{footnote}]{\footnotemark[#1]}

\author{Saisubramaniam Gopalakrishnan\thanks{Equal Contribution.} \and Pranshu Ranjan Singh\samethanks[1] \and Yasin Yazici \and Chuan-Sheng Foo \and Vijay Chandrasekhar\thanks{Contribution in 2019.} \and ArulMurugan Ambikapathi}
\authorrunning{Saisubramaniam et al.}
%
\institute{Institute for Infocomm Research, A*STAR}

\maketitle

\begin{abstract}
Utilization of classification latent space information for downstream reconstruction and generation is an intriguing and a relatively unexplored area. In general, discriminative representations are rich in class specific features but are too sparse for reconstruction, whereas, in autoencoders the representations are dense but has limited indistinguishable class specific features, making it less suitable for classification. In this work, we propose a discriminative modelling framework that employs manipulated supervised latent representations to reconstruct and generate new samples belonging to a given class. Unlike generative modelling approaches such as GANs and VAEs that aim to model the data manifold distribution, Representation based Generations (ReGene) directly represents the given data manifold in the classification space. Such supervised representations, under certain constraints, allow for reconstructions and controlled generations using an appropriate decoder without enforcing any prior distribution. Theoretically, given a class, we show that these representations when smartly manipulated using convex combinations retain the same class label. Furthermore, they also lead to novel generation of visually realistic images. Extensive experiments on datasets of varying resolutions demonstrate that ReGene has higher classification accuracy than existing conditional generative models while being competitive in terms of FID.
\end{abstract}

\blfootnote{Full published version: \href{https://doi.org/10.1016/j.neucom.2021.10.090}{https://doi.org/10.1016/j.neucom.2021.10.090}}

\section{Introduction}
\begin{figure}[htpb]
	\centering
	\includegraphics[width=1.0\linewidth]{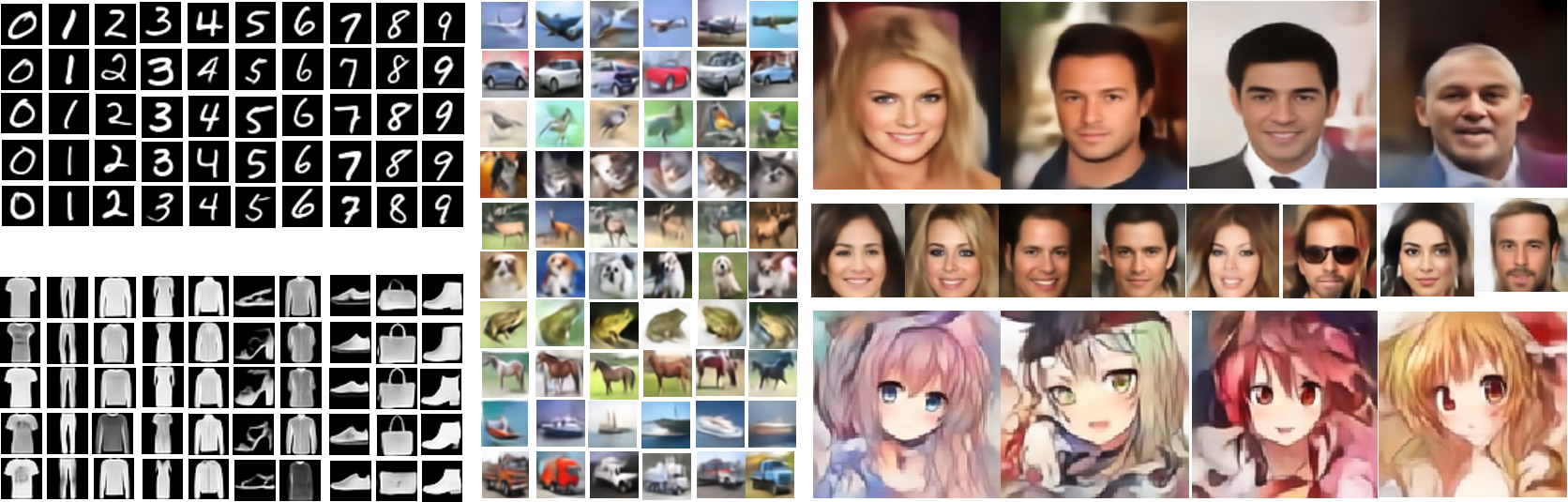}
	\caption{Image generations of varying resolutions - MNIST, Fashion MNIST (28x28), CIFAR-10 (32x32), CelebA (64x64 and 128x128), and Anime (128x128).}
	\label{fig:fig1}
\end{figure}
Image classification is one of the major areas of computer vision that is being tremendously revolutionized by Deep Learning (DL). Since the work on AlexNet \cite{KrizhevskySH2017alexnet}, there have been significant contributions including ResNet \cite{he2016deep}, DenseNet \cite{huang2017densely}, InceptionNet \cite{szegedy2016rethinking} to name a few, that have pushed the boundaries in image classification due to their increased accuracy. The latent space representations of these classifiers hold a rich set of distinguishing features pertaining to each class, inspiring us to explore the utilization of such representations to further generate different valid representations belonging to a given class. Such generated representations can then be used for new sample / image generations, belonging to a given class. Recently, DL based generative modeling has manifested itself as a fascinating and promising research area for image generation. Recent advancements in generative modeling include Autoencoders based generative models \cite{KingmaW2013autoencoding, tolstikhin2018wasserstein, dai2018diagnosing,  sohn2015learning, ZhangZLBP2019perceptual}, GAN based methods \cite{GAN_Original_2014, DCGAN_Radford_ICLR2016, arjovsky2017wasserstein, gulrajani2017improved, BerthelotSM2017began}, and Flow based methods \cite{DinhSB2017realnvp, kingma2018glow}. A plethora of algorithms based on these ideas (and their variants) have revolutionized and redefined the notion of image generations. As much as the interest for image generations have been shown towards generative models, to the best of our knowledge, using discriminative approaches for the possibility of generations from classification latent space is comparatively less explored. Alternatively, similar to autoencoder and its variants being subjected to the downstream task of classification, we are interested in training a classifier first and later exploit the learnt classification latent space for downstream image reconstruction and generation.

\begin{figure}[t]
	\centering
	\includegraphics[width=1.0\textwidth]{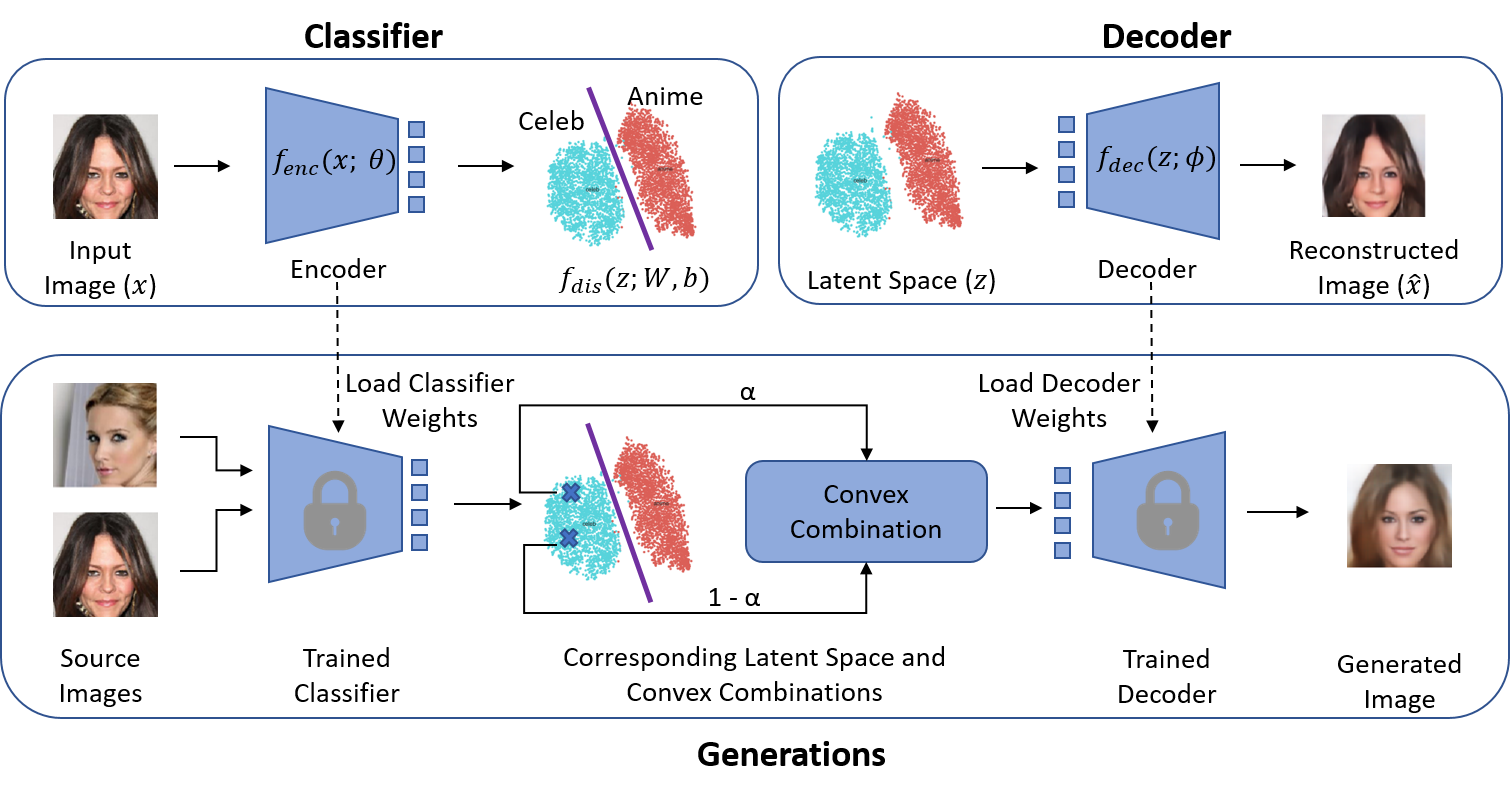}
	\caption{Image generation using ReGene: Top blocks show classifier (for supervised latent space representations) and decoder (for image reconstruction) to be modeled. Bottom row depicts the image generation procedure using the trained classifier and decoder.}
	\label{fig:overall_regene1}
\end{figure}

To begin with, few natural questions might arise - Can latent space representations learned by a network trained exclusively for classification task be reused for another downstream task such as reconstruction and / or  generation? Can we substitute the need for a latent space prior (as in VAEs/GANs) with a property from classification latent space? In this work, we endeavor to address the above questions with an aim to create additional latent space representations from a set of representations belonging to samples of a given class, that are still guaranteed to belong to that class. For such obtained latent space representations, an appropriately designed decoder can result in generation of visually meaningful samples belonging to the given class of interest. In this regard, we propose a framework, namely Representation based Generations (ReGene), that investigates the above thoughts and demonstrates the feasibility of such generations from classification latent space representations. 
The main contributions here are as follows: (i) We theoretically show that classifier latent space can be smartly interpolated using convex combinations, to yield new sample representations within the manifold. (ii) We discuss how to select good latent space representations that are sufficient for reconstructing respective images through the design of an appropriate decoder using a combination of loss functions. (iii) Finally, we demonstrate how convex combinations of latent representations ($z$) of images ($X$) belonging to a class can lead to realistic and meaningful generations of new image samples belonging to the same class, using a decoder network exhibiting good generalization capability (i.e., $p(X|z)$). The overall ReGene framework is built based on discriminative modelling approach and capable of generating new images (from convexly combined latent space representations) belonging to a class, is depicted in Figure \ref{fig:overall_regene1}, and the associated details are presented in the ensuing sections. 
\section{Background and Related Work}
In this section, we discuss the literature pertaining to latent space representations and image generations inline with the ReGene framework.
\subsection{A Study on Latent Spaces: Autoencoder Vs Classification Latent Space}
There are several interesting works on the exploration of latent spaces for supervised (classification latent space) and unsupervised (autoencoder) tasks \cite{ng2011sparse, Goodfellow-et-al-2016}. Autoencoders learn compressed latent space representations but can also produce realistic interpolated images by imposing additional constraints. For instance, \cite{Goodfellow_Understanding_Interpol_ICLR2019} proposed an adversarial regularizer on the generated samples from autoencoder latent space interpolations and \cite{Sainburg_Adv_Interpol_AE_2018} trained adversarially on latent space interpolations. Autoencoder latent space representations can be utilized for downstream tasks by adding relevant regularizers \cite{MakhzaniF2014ksparse, Bengio_BetterMixing_2013, Goodfellow_Understanding_Interpol_ICLR2019}. Autoencoder latent space for generations of image samples has been exploited to great extent, but classification latent space based image generations is unexplored in computer vision community.  

\begin{figure}[htbp]
\centering
\subfloat[]{
    \includegraphics[width=0.48\linewidth]{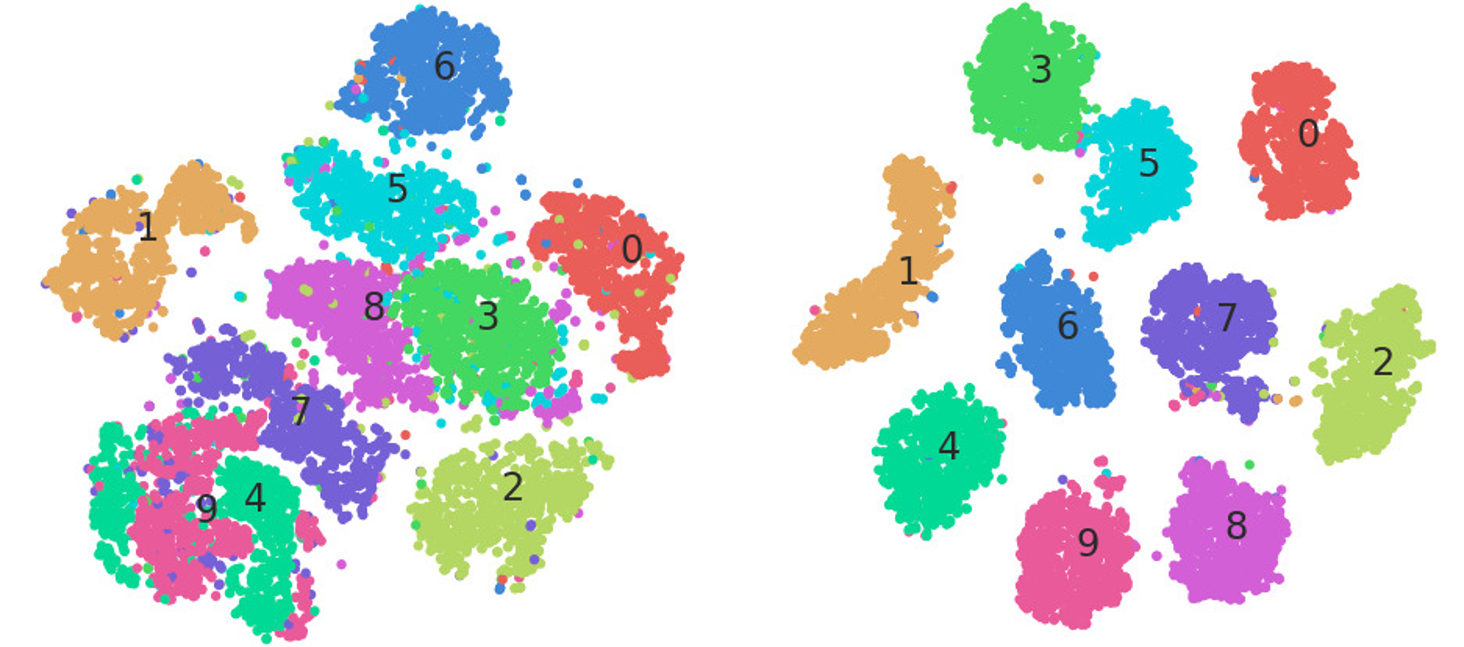}
    \label{mnist-autoencoder-classifier-latent-space-comparison}
}
\subfloat[]{
    \includegraphics[width=0.48\linewidth]{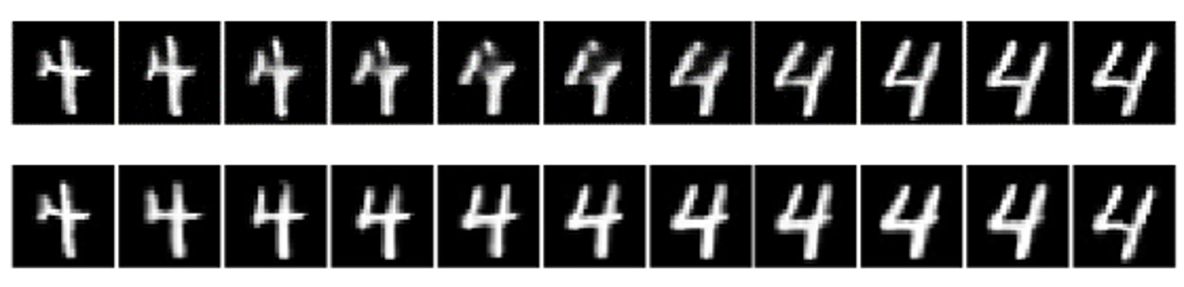}
    \label{mnist-autoencoder-classifier-latent-space-comparison-transition-digit-4}
}
\caption{(a) MNIST latent representation from a trained Autoencoder (left) and Classifier (right) (b) MNIST latent representation from a trained Autoencoder (top) and Classifier (bottom)}
\end{figure}

A comparison in interpolations between Autoencoder and Classifier latent spaces using MNIST dataset can be considered as a baseline illustration. The t-SNE plots for the autoencoder and classifier latent space for MNIST dataset are shown in Fig. \ref{mnist-autoencoder-classifier-latent-space-comparison}. Fig. \ref{mnist-autoencoder-classifier-latent-space-comparison} (left) shows that the linear interpolation of points belonging to a class do not always belong to the same class. For instance, the line between two points belonging to the cluster 4 (green) passes through the cluster 9. Therefore, the generated images based on such combinations are not realistic and meaningful. One such illustration is shown in Fig. \ref{mnist-autoencoder-classifier-latent-space-comparison-transition-digit-4} (top) where the transition can be seen blurry as it moves from left (a given true sample) to right (another true sample). On the other hand, for the same two true samples, Fig. \ref{mnist-autoencoder-classifier-latent-space-comparison-transition-digit-4} (bottom) shows the classification latent space based transition, where one can observe a smooth transition within intermediate samples still preserving the class information (in this case ``4"). This illustrates that the classifier space will be apt for generating new meaningful samples that still belongs to the same class of interest.


\subsection{Generative Modeling - gaining popularity for Image Generations}
GANs and VAEs lead the list of modern generative learning approaches and have shown tremendous progress in recent years for generation of new images. Vanilla versions of both the approaches model the data distribution in an unsupervised fashion (unlabelled). 
VAEs enforce a prior distribution to control latent representations, whereas, GANs do not have an encoder to provide latent representations and require random noise from a prior distribution. GANs tend to suffer from issues such as mode collapse/mode drop, training instability, etc., which have been addressed in \cite{SalimansGZCRCC2016improved, DCGAN_Radford_ICLR2016}. VAEs may produce blurry images and recent methods address these issues and shown sharp images \cite{ZhangZLBP2019perceptual, ali2019vqvae2}. Conditional GANs such as \cite{mirza2014conditional, isola2017image} and Conditional VAEs \cite{sohn2015learning, IvanovFV2019variational} learn conditional data distributions, the condition being class labels or other images.  
Generating high-resolution and quality images is a  challenge for both GANs and VAE. BigGAN and Progressive GAN are recent works that address this problem. \cite{BrockDS19biggan, Prog_GAN_2018}. VQ-VAE, leveraging on discrete latent embedding also demonstrates comparable performance to state-of-the-art GAN models \cite{ali2019vqvae2}. Though GANs and VAEs are the most studied generative modeling approaches, it should be emphasized that ReGene is a discriminative modeling framework and takes a different approach towards representative modeling based generation.


\section{Representation Based Generations (ReGene)}

The prime focus of ReGene framework is to capture the distribution of the data belonging to a class, so that new images from that class can be generated. To achieve this, the ReGene framework involves three parts (as shown in Figure \ref{fig:overall_regene1}): (i) Classifier: Encoder for supervised latent space representation, (ii) Decoder: Image reconstruction and generation from the set of latent representations, and (iii) Convex analysis based manipulation of latent space representation. In this section, we derive the mathematical formulations that theoretically explains the framework's ability to capture the data distributions. Let $\bm{X} = \{x^{(i)}\}_{i=1}^{m}$, $\bm{y} = \{y^{(i)}\}_{i=1}^{m}$ be $m$ i.i.d. data samples and the corresponding class labels, respectively.  


\subsection{Classifier: Encoder - Image Space to Latent Space}
The purpose of the encoder here is to find appropriate latent space representations to classify the data samples according to their respective class labels (the well-known classification problem). For sake of clarity, let $f_{cls}(x;\theta, W, b)$ be the classifier that maps the data sample $x$ to class label $y$. This classifier can be written as composition of two functions: \newline
(i) Encoder $f_{enc}(x;\theta)$, which maps the data sample $x$ to latent space representation $z$; \newline 
(ii) Discriminator $f_{dis}(z;W, b)$, which maps  the latent space representation $z$ to the class label $y$. \newline
Hence, 
\begin{equation}
f_{cls}(x;\theta, W, b) = f_{dis}(f_{enc}(x;\theta);W, b).    
\end{equation}

Let $p(y|x)$ be the probability of classifying $x$ according to class labels $y$ for each data sample. Then, the overall classifier likelihood function can be defined as
\begin{equation} \label{loss_eqn11}
L(\theta, W, b) = \mathop{\mathbb{E}}_{x, y \sim p_{x, y}}[\log p(y^{(i)}|x^{(i)} ; \theta, W, b)].
\end{equation}

By jointly optimizing the classifier using equation \eqref{loss_eqn11}, the obtained model parameter estimates $\theta^{*}$ yield the optimal latent space representation $z^{*}$, which is defined as:  
\begin{equation} \label{latent_code}
z^{*} = f_{enc}(x;\theta = \theta^{*}).
\end{equation}
These latent space representations will then be used to reconstruct / generate images, as discussed next.

\subsection{Decoder: Latent space to Image space Reconstruction}

Let $f_{dec}(z;\phi)$ be the decoder, which maps the optimal latent space representation $z^*$ to data sample $x$. Let $p(x|z^{*})$ be the conditional probability of obtaining the sample $x$ given the latent space representation $z^{*}$ via the process of generalized reconstruction. Then the overall decoder likelihood function can be defined as
\begin{equation} \label{loss_eqn1}
L(\phi) = \mathop{\mathbb{E}}_{z^{*}, x \sim p_{z^{*}, x}}[\log p(x^{(i)}|{z^{*}}^{(i)} ; \phi)].
\end{equation}
A decoder designed by optimizing the above function will model $p(x|z^{*})$ and hence can be used for reconstruction and generation, provided the $z$ used for the generation still falls within the distribution of $p(x|z^{*})$. It should be emphasized that in ReGene framework a single decoder is simultaneously trained for all the classes. In other words, the decoder aims to invert the classification latent space representations and therefore expected to preserve class information. 

\subsection{Convex combinations of latent space representations}
An optimal decoder following the above discussions ensures that the classification space is sufficient for effective reconstruction of images. Now, we analyze the latent space representations of the members of a class. Particularly, we show that the classification latent space is manipulative and the convex combination of two or more latent space representations is still classifiable as the same class. 

\begin{lemma}
\label{lemma1}
Considering a binary classification scenario: Let $A =\{{\bf a}_1,\ldots,{\bf a}_{n_A}\} \subset \mathbb{R}^{n}$ be an $n$ dimensional feature space representation of a class $A$ that contains ${n_A}$ elements. And let $B =\{{\bf b}_1,\ldots,{\bf b}_{n_B}\} \subset \mathbb{R}^{n}$ be an $n$ dimensional feature space representation of a class $B$ that contains ${n_B}$ elements. If there exists a separating hyperplane ${\bf W}^T{\bf z} + b = 0$, that separates these two classes such that:
\begin{align}\label{eq:hyperplane_1}
{\bf W}^T{\bf z} + b &< 0, \forall {\bf z} \in A \\
{\bf W}^T{\bf z} + b &> 0, \forall {\bf z} \in B.
\end{align} 
Then, it is true that 
\begin{align}\label{eq:condn_A_1}
{\bf W}^T{\bf z} + b &< 0, \forall {\bf z} \in {\rm conv}\{{\bf a}_1,\ldots,{\bf a}_{n_A}\}
\end{align}
\begin{align}\label{eq:condn_B_2}
{\bf W}^T{\bf z} + b &> 0, \forall {\bf z} \in {\rm conv}\{{\bf b}_1,\ldots,{\bf b}_{n_B}\},
\end{align}
where {\it convex hull} of $\{ {\bf a}_1,\ldots, {\bf a}_m \}$ $
\subset \mathbb{R}^n$ is defined as
\begin{equation}\label{eq:defn_1}
{\rm conv}\{ {\bf a}_1,\ldots, {\bf a}_m \} = \bigg\{ {\bf z} = \sum_{i=1}^m \theta_i {\bf a}_i  \bigg|  {\bf 1}_m^T\bm{\theta}=1, {\bm \theta} \succeq {\bm 0}\bigg\}.
\end{equation}
\end{lemma}
\begin{proof}
Let us begin by proving \eqref{eq:condn_A_1}. Let ${\bm \alpha} \in {\rm conv}\{{\bf a}_1,\ldots,{\bf a}_{n_A}\} $.
Then, ${\bf W}^T{\bm \alpha} + b $ can be written as:
\begin{align}
\begin{split}
=~& {\bf W}^T(\theta_1{\bf a}_1+\theta_2{\bf a}_2+\cdots+\theta_{n_A}{\bf a}_{n_A}) + b,\\ 
 ~&{\rm where} ~\theta_i \geq 0,~i=1,\ldots,n_A, {\bf 1}_{n_A}^T\bm{\theta}=1~  ({\rm by~\eqref{eq:defn_1}})
\end{split} \\
\begin{split}
=~&\theta_1{\bf W}^T{\bf a}_1 + \theta_2{\bf W}^T{\bf a}_2 + \cdots + \theta_{n_A}{\bf W}^T{\bf a}_{n_A}\\ ~&+ b{\bf 1}_{n_A}^T\bm{\theta}, ~({\rm as~} {\bf 1}_{n_A}^T\bm{\theta} = 1)
\end{split}\\
\begin{split}
=~&\theta_1{\bf W}^T{\bf a}_1 + \theta_2{\bf W}^T{\bf a}_2 + \cdots + \theta_{n_A}{\bf W}^T{\bf a}_{n_A}\\ ~&+ (\theta_1{b} + \theta_2{b} + \cdots + \theta_{n_A}{b})
\end{split}\\
\begin{split}
=~&\theta_1({\bf W}^T{\bf a}_1 + b) + \theta_2({\bf W}^T{\bf a}_2 + b) + \cdots\\ ~&+ \theta_{n_A}({\bf W}^T{\bf a}_{n_A} + b)\end{split}\\
<~&0 ~({\rm by~\eqref{eq:hyperplane_1}})
\end{align}
Following similar steps, \eqref{eq:condn_B_2} can also be proved. ~~~~~~~~~~~~~~~~~~~~~~~~~~~~~~~~~~~~~~ 
\end{proof}

\subsection{Generations in Image Space}

Consequently, as the decoder is trained to capture the distribution of x through a finite set of latent space representations $z^*$, it can be shown that it can generate meaningful new images for each and every element belonging to the convex hull of the set of $z^*$, corresponding to a given class. The generations in image space are obtained by performing following two steps:\newline
(i) Obtain generations in latent space for a given class via convex combinations of any $n$ number of latent space representations ($n~\geq~2$), belonging to that class.\newline
(ii) Trigger the trained decoder with the newly obtained latent space representations to get corresponding image space generations.

The latent space points obtained from step (i) lie within the given class (see Lemma \ref{lemma1}). Though we do not enforce a prior and cannot take advantage of infinite random generations arising from the prior, we can, however, achieve countably infinite number of generations by varying the sample count participating in the generation process and their convex combination ratio. The validity of image space generation obtained by the decoder in step (ii) is presented in Theorem \ref{theorem3}. From \cite{Vladik2018whymixture} we have the following result, which is used in the proof of Theorem \ref{theorem3}.
\begin{theorem}
\label{prop2}
Given two random variables $\alpha_{1}$ and $\alpha_{2}$, with probability density functions $\rho_{1}(x)$ and $\rho_{2}(x)$ respectively, the probability density function $\rho(x)$ of the mixture obtained by choosing $\alpha_{1}$ with probability $w$ and $\alpha_{2}$ with remaining probability $1-w$ is a convex combination of probability density functions of original random variables.
$\rho(x)$ = $f(\rho_{1}(x), \rho_{2}(x)$) ~= ~$w . \rho_{1}(x) + (1 - w) . \rho_{2}(x)$
\end{theorem}

\begin{theorem}
\label{theorem3}
For an ideal decoder, 
the convex combinations of latent space representations per class yields image representations belonging to the original image space distribution.   
\end{theorem}
\begin{proof}
Let the classification latent space obtained be $z^{*}$,
\begin{equation} \label{latent_code_2}
z^{*} = f_{enc}(x;\theta = \theta^{*}).
\end{equation}
Let $f_{dec}(z^{*};\phi)$ be the decoder, that learns a mapping from the latent space distribution to image space distribution and $z^{*}|y_{i}$ be the latent space distribution for class $y_{i}$. \newline
The latent representations per class,
\begin{equation}
z^{*}|y_{i} \sim p(z|y_{i})
\end{equation}
For the mixture of two latent representations from a given class chosen with probabilities $\alpha$ and $1 - \alpha$ respectively, the probability density is given by,
\begin{align}
\begin{split}
f(p(z|y_{i}), p(z|y_{i})) =~& \alpha . p(z|y_{i}) + (1 - \alpha) . p(z|y_{i}) \\
 (&{\rm by~Theorem~\ref{prop2}}) \\
 =~& p(z|y_{i})
\end{split}
\end{align}
Hence, the mixture of two latent representations for a given class preserves the latent space distribution for that class. So, the decoder, $f_{dec}(z^{*};\phi)$ is able to map this latent space representation to a sample in the image space. Since the output of decoder belongs to the original image space distribution, the newly generated image sample belongs to the original image space distribution. ~~~~~~
\end{proof}
\section{Experimental Setup}
In this section we discuss the experimental procedures 
and provide a high level overview of the classifier and decoder network architectures employed here for analyzing ReGene framework. We also discuss the evaluation metrics for performance comparison with state-of-the-art methods for generation.
\subsection{Datasets}
\label{datasets}
Four standard datasets for image generation tasks - MNIST \cite{lecun2010mnist}, Fashion MNIST \cite{DBLP:journals/corr/abs-1708-07747}, CIFAR-10 \cite{Krizhevsky09learningmultiple} and CelebA  \cite{LiuLWT15} are selected. MNIST and Fashion MNIST are relatively easier due to smaller image dimensions (28x28 px), grayscale images, perfect alignment, and absence of background. CelebA is chosen to demonstrate the generation of real-world, relatively higher-dimensional color images. To be consistent with other reported results in the literature, we followed the same pre-processing steps provided in \cite{tolstikhin2018wasserstein} and \cite{dai2018diagnosing} by taking 140x140 center crops and resizing the image to 64x64 px resolution. \footnote{It should be noted that existing works in literature have used different versions of CelebA dataset with respect to pre-processing. Additional details pertaining to this are provided in supplementary material.} To obtain latent space representation for CelebA, we introduce Anime \cite{danbooru2018} dataset as the complementary class.
15K random samples from both CelebA and Anime were chosen for training the classifier. For the decoder, we trained CelebA separately using the entirety of 200K samples. Additionally, we also trained CelebA and Anime for 128x128 px resolution without additional cropping other than provided in the original aligned dataset. CIFAR-10 (32x32 px) is an interesting dataset of choice since it presents a multiclass scenario with high intraclass variations and unaligned images with mixed backgrounds.
For quantitative evaluation on the datasets, we directly state the values of the evaluation metrics (as reported in the respective literature), to have fair comparisons with other existing approaches.

\subsection{Network Architectures}
It should be noted that the prime focus of this work is on the theory and principle of the representation framework that suits the downstream task of generation, and not intended to propose an optimal architecture for classifier or decoder. Therefore, here we demonstrate our approach using simple feed-forward neural networks. Both classifier and decoder networks follow VGG-style architectures of varying depths and multiple blocks, with each major block having Conv2D for classifier / Conv2DTranspose for decoder, with BatchNorm, LeakyReLU, followed by a MaxPooling for classifier / UpSampling for decoder layer. Cross-entropy is adopted as the loss function for classification since it aims to maximize the capture of information sufficient enough to separate the classes. For the decoder, we employ a combination of three weighted loss functions: Mean Absolute Error (MAE), Structural Similarity Index (SSIM) \cite{WangBSS2004ssim} and Perceptual Loss \cite{perceptual_loss}.
The ratio of weights are dataset dependent;
however the general heuristic is to allow initial training
epochs to have high weights assigned to SSIM and Perceptual
Loss to first reconstruct the global outline and texture
of the image. In later epochs, those weights are gradually
reduced to focus more on pixel wise error. The MAE is
weighted relatively high throughout the training. Network weights are optimized using Adam optimizer. All experiments were run on a workstation with the following specifications: 64 GB RAM, Intel i9-9900K CPU and 2x NVIDIA GeForce RTX 2080 Ti. Detailed network architectures and training details are provided in the supplementary material.  


\label{cifar10-case-study}
Next, we discuss the architecture selection for a good latent space representation. For a complex dataset with high intra-class and background variations, learning proper class-specific features become difficult. Taking CIFAR-10 as an example, we examine how to achieve trade-off between the ideal classification and reconstruction. 
Generally, discriminative classification models often yield highly sparse latent representations that sacrifice reconstruction quality, implying even a decoder with deep architecture will fail if it does not receive sufficient latent space information.
With the intention of finding good latent space representations that achieve a balance between classification and reconstruction / generation, a hyper-parameter search was conducted for a suitable architecture in terms of depth (layers), latent space dimension size and relative non-sparsity in the latent feature space. 
It was observed that although architectures with more layers achieve higher accuracy, the latent code gradually appears sparser, which affects the decoder. 
The classification network architecture having latent space $2048$ and single convolution per layer was chosen as it achieves best decoder MAE. Again, it should be emphasized that the classifier (encoder) and the decoder used here are not necessarily the best / optimal ones. However, they are sufficient to demonstrate the efficacy of ReGene framework. A detailed analysis of this case study also including DenseNet is provided in the supplementary material. Similar approach was adopted to find the right choice of classifier and decoder combination for other datasets.

\subsection{Evaluation Metrics}
To quantitatively evaluate the generations, we report their Classification Accuracy and Fréchet Inception Distance (FID) \cite{heusel2017gans}. The classification accuracy validates the claim from Lemma \ref{lemma1} from the perspective of a neutral classifier (ResNet56v1 architecture), trained from scratch on the original dataset corresponding to the generations. On the other hand, FID which is used to quantify the generations is also susceptible to the decoder noise, which is further discussed in Section \ref{FIDSECTION}. For qualitative evaluation, we compare generations with nearest neighbors in the original training image space, their reconstructions, and with other generations created.



\section{Results and Discussion}

\subsection{Validation of Lemma 1 and Theorem \ref{theorem3}}
\begin{figure}[htbp]
\centering
\subfloat[]{
    \includegraphics[width=0.45\linewidth]{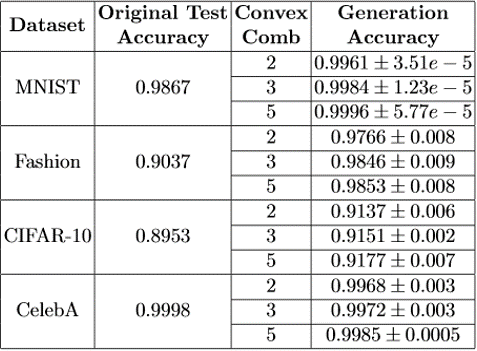}
    \label{table::tab4}
}
\subfloat[]{
    \includegraphics[width=0.51\linewidth]{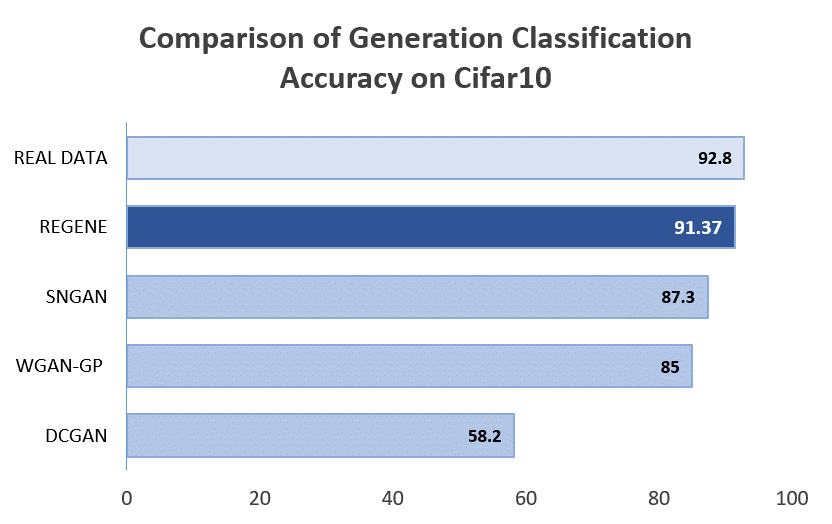}
    \label{fig::classification_accuracy_cifar10}
}
\caption{(a) Generation Classification Accuracy on Neutral Classifier for different dataset generations with convex combination - 2, 3 and 5. (b) Comparison on Cifar10 dataset. ReGene outperforms other conditional generative approaches in terms of generation classification accuracy on Cifar10. Scores for baseline GANs are reported from \cite{ShmelkovSA2018how}.}
\end{figure}

The validity of Lemma \ref{lemma1} can be inferred from the Table provided in Figure \ref{table::tab4}, wherein the generation classification accuracies from the viewpoint of a neutral classifier ResNet56v1 are all high ($\geq$ 0.9). It is observed that convex combinations with more samples tend to provide generations with overall higher accuracy. Figure \ref{fig::classification_accuracy_cifar10} compares the classification accuracy score of ReGene for Cifar-1o dataset, with those of the reported baseline conditional generative models. It is evident from Figure \ref{fig::classification_accuracy_cifar10} that ReGene generations are better classifiable by a neutral classifier. Although a perfectly practical validation for Theorem \ref{theorem3} can be obtained only from an ideal decoder, the comparative scores of FID (reported and discussed in Section \ref{FIDSECTION}) help to substantiate that the decoder indeed closely learns the mapping from latent space to image space distribution and that the generations are not far from the original image space distribution.

\begin{figure}[htbp]
    \centering
    \includegraphics[width=1.0\linewidth]{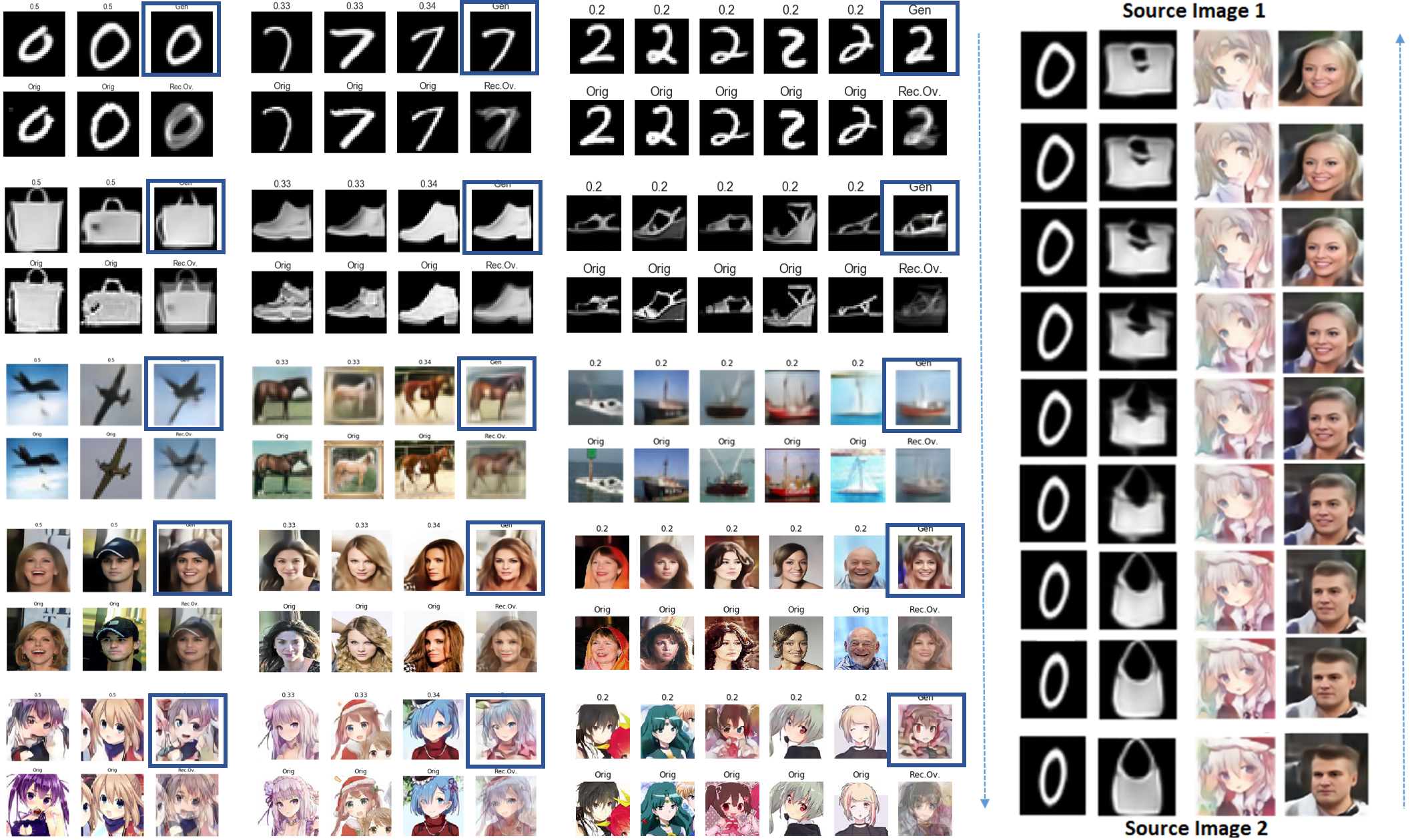}
    \caption{Left: Generations (highlighted in the blue square) produced from convex combination of different samples - n=2,3,5 with combination ratio of 1/n. Right: Latent Space Interpolation transitions between two source images  (top and bottom of each column) of digit 0, bag, anime and celeb, shown vertically for varying convex combination ratios (0.1-0.9)  (All images are shown in same scale due to space constraints though they have different resolutions).}
    \label{all_generations_cc_collage}
\end{figure}

\begin{table}
\scriptsize
\centering
\caption{Comparison of FID scores. For benchmarking purposes, results indicated by $^\dagger$ are taken from \cite{dai2018diagnosing}, $^\diamond$ from \cite{ZhangZLBP2019perceptual}.  
The blanks indicate that the values are not reported in the respective papers. ReGeNe generations are obtained by performing convex combinations of 2 samples, experiments repeated over 3 separate runs. The last row represents FID scores computed between dataset reconstruction and generations. For each dataset the top 2 performances are highlighted. 
}
\begin{tabular}{|c|c|c|c|c|}
\hline
\textbf{Method} & \textbf{MNIST} & \textbf{Fashion} & \textbf{CIFAR-10} & \textbf{CelebA} \\ \hline
NS GAN$^\dagger$ & $\bf{6.8 \pm 0.5}$ & $26.5 \pm 1.6$ & $58.5 \pm 1.9$ & $55.0 \pm 3.3$ \\ \hline 
LSGAN$^\dagger$ & $7.8 \pm 0.6$ & $30.7 \pm 2.2$ & $87.1 \pm 47.5$ & $53.9 \pm 2.8$ \\ \hline 
WGAN GP$^\dagger$ & $20.3 \pm 5.0$ & $24.5 \pm 2.1$ & $55.8 \pm 0.9$ & $30.3 \pm 1.0$ \\ \hline
BEGAN$^\dagger$ & $13.1 \pm 1.0$ & $\bf{22.9 \pm 0.9}$ & $71.4 \pm 1.6$ & $38.9 \pm 0.9$ \\ \hline
VAE$^\diamond$ & $19.21$ & $-$ & $106.37$ & $48.12$ \\ \hline
CV-VAE$^\diamond$ & $33.79$ & $-$ & $94.75$ & $48.87$ \\ \hline
WAE$^\diamond$ & $20.42$ & $-$ & $117.44$ & $53.67$ \\ \hline
RAE-SN$^\diamond$ & $19.67$ & $-$ & $84.25$ & $44.74$ \\ \hline
LVPGA$^\diamond$ & $11.4 \pm 0.25$ & $-$ & $\bf{52.9 \pm 0.89}$ & $\bf{13.8 \pm 0.20}$ \\ \hline
2-Stage VAE$^\dagger$ & $12.6 \pm 1.5$ & $29.3 \pm 1.0$ & $72.9 \pm 0.9$ & $44.4 \pm 0.7$ \\ \hline
\hline 
ReGene (\emph{Org vs Gen}) & $16.3 \pm 0.06$ & $40.9 \pm 0.16$ & $59.7 \pm 0.24$ & $48.3 \pm 0.32$ \\ \hline
ReGene (\emph{Org vs Recon}) & $13.9$ & $34.2$ & $52.7$ & $32.9$ \\ \hline
ReGene (\emph{Recon vs Gen}) & $\bf{3.03 \pm 0.02}$ & $\bf{4.03 \pm 0.03}$ & $\bf{6.30 \pm 0.03}$ & $\bf{10.82 \pm 0.04}$ \\ \hline
\end{tabular}
\label{table::tab3}
\end{table}
\subsection{Generations via Convex Combination}
We discuss the generations of new samples from the convex combination of multiple existing samples ($n\geq 2$) in the latent space. By theory, we can have countably infinite samples participating in the generation process. For practical purposes, due to classification and decoder errors, we apply a threshold that serves as filtering criteria to select the best generations. We apply a two-step judging criteria to decide whether a sample from the combination is fit to be considered as a new generation: (i) On obtaining the new latent space representation, pass to the classifier and allow samples that have class confidence score above a certain threshold, (ii) Such sample representations once decoded to image (via decoder) are again passed to the classifier to double check whether the new image is also of high class confidence. General threshold adopted is the average class confidence per class over the test/holdout samples. In Figure \ref{all_generations_cc_collage}, we show generations from combinations n=2,3,5 and the convex combination ratio=1/n. Each row presents generations (highlighted in the blue square) from datasets selected in Section \ref{datasets}. Below each of the generated images (highlighted blue square) we have shown the spatial overlapping of the source images to visually observe the novelty and quality of the generated samples. Each column shows samples from different number of samples participating in the combinations. The generations - Digit 0, Bag, Anime, and Celeb transitions are produced by changing the convex combination ratios (0.1 - 0.9) in the latent space. The images reveal the generative ability of ReGene (for different $n$ and in various datasets) and the smooth image transitions (for $n=2$).{\footnote{Owing to page limitations, extensive image generations for various datasets are provided in the supplementary material.}}
\subsection{Finding the Closest Match}
\begin{figure}[t]
\centering
\subfloat[]{
    \includegraphics[width=0.62\linewidth]{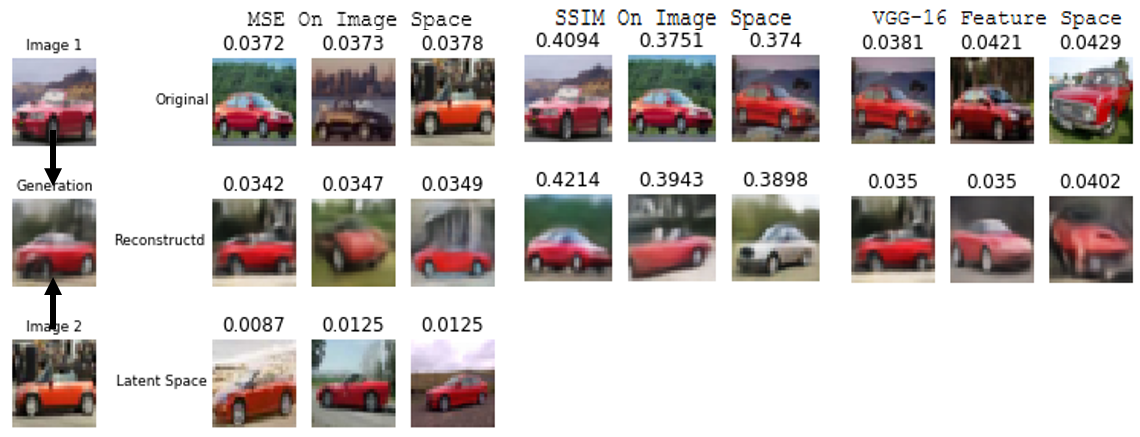}
    \label{celeb_genertion_original_recon_comparison}
}
\subfloat[]{
    \includegraphics[width=0.33\linewidth]{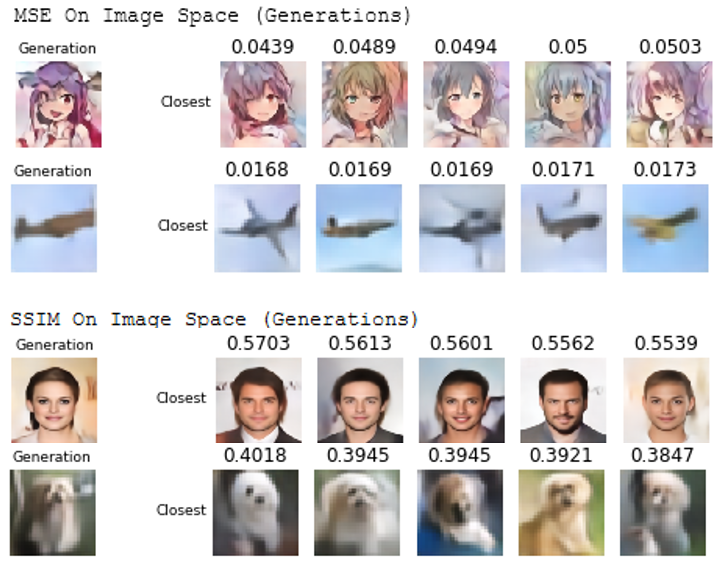}
    \label{celeb_genertion_comparison_itself}
}
\caption{(a) Comparison of generation among the training set samples - Showing Top 3 Original and Reconstructed using MSE, SSIM in Image Space and VGG-16 Latent Feature Space. (b) Comparison of generation among other generations - First two rows show top 5 closest matches in increasing order of squared error in image space; bottom two rows show top 5 closest match in terms of decreasing order of SSIM in image space.}
\end{figure}

To demonstrate that the decoder does not produce generations that simply mimic existing samples, and to show it is truly robust to mode collapse, we show qualitative samples comparing generations with samples from the training set (Figure \ref{celeb_genertion_original_recon_comparison}). We take two existing images, generate a new image and compare it against other existing images from the training dataset, by closest match in (i)  increasing order of squared error in image and latent space representations, (ii) decreasing order of SSIM in image space, and (iii) increasing order of squared error on latent features passed through Imagenet pre-trained VGG-16 network. We compare with our own created generations set in Figure \ref{celeb_genertion_comparison_itself}.
\begin{figure}[t]
\centering
\subfloat[]{
    \includegraphics[width=0.48\linewidth]{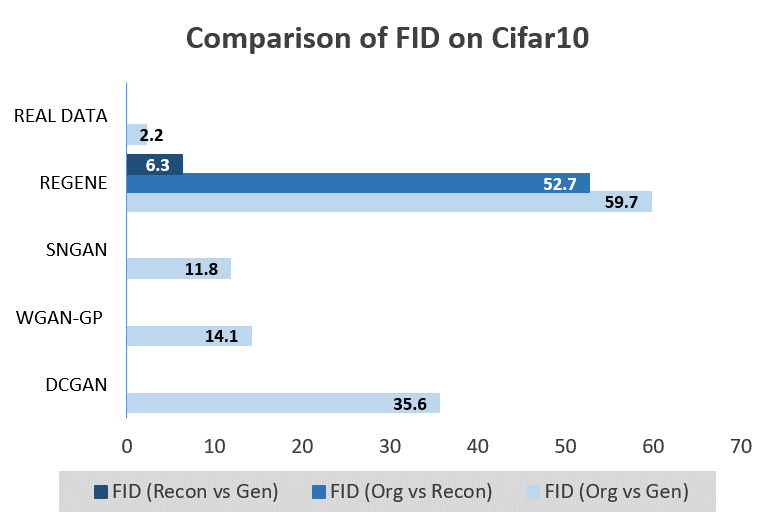}
    \label{cifar_fid_chart}
}
\subfloat[]{
    \includegraphics[width=0.42\linewidth]{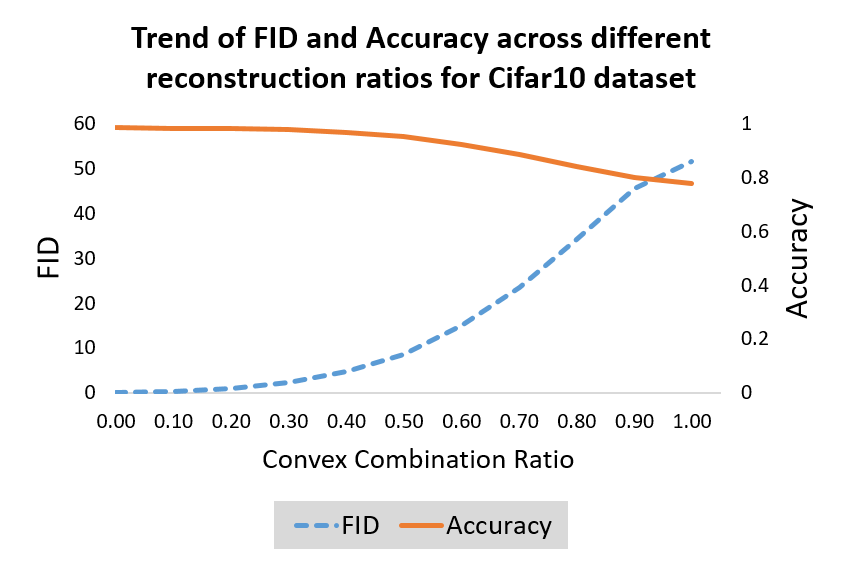}
    \label{cifar_reconstruction_ratio_fid_accuracy_trend}
}
\vspace{-0.35cm}
\caption{Left: Comparison of FID on the Cifar10 dataset - The FID score for Recon vs Gen (6.3) is the best among all. However, it should be noted that (Org vs Gen) FID score (59.7) is higher than other conditional generative models. However, (i) ReGene was never optimized directly for generations, (ii) the (Org vs Recon) FID score (52.7) is also high, indicating the reconstruction error should be accounted for during comparison. Right: Trend of FID score (blue) vs Accuracy (orange) for different convex combination ratios of original:reconstructed samples. Higher FID score is observed as the ratio tends towards reconstructed samples, caused primarily by the decoder error.}
\vspace{-0.4cm}
\end{figure}

\subsection{Comparing FID scores of Reconstruction vs Generation}
\label{FIDSECTION}
The decoder of ReGene serves for both reconstruction and generation. Unlike GANs/Conditional GANs, ReGene does not employ adversarial training. The FID scores obtained using ReGene generations for different datasets, and those obtained by different generative approaches (as reported in the literature), are summarized in Table \ref{table::tab3}. It is important to note that the FID scores computed between Org vs Recon are high due to low level decoder noise. This does not affect generation qualitatively (as can be witnessed from images in Figure \ref{all_generations_cc_collage}), but it impacts FID score significantly. As FID score is susceptible to such small decoder noise, it favor methods trained in adversarial fashion. The FID scores between Recon vs Gen provide a better comparison since they both (reconstruction and generation from same decoder) take into account the same decoder noise, and it beats the state-of-the-art methods. Similar observation has been reported in \cite{ali2019vqvae2}. Figure \ref{cifar_fid_chart} compares the generative performance exclusively for Cifar-10 database, where again the FID score for Recon vs Gen beats the state-of-the-art methods. In Figure \ref{cifar_reconstruction_ratio_fid_accuracy_trend}, we plot FID score vs accuracy for different convex combination ratios of original-reconstructed versions of the Cifar-10 samples. Initially, when all original training images are presented, the accuracy reaches 1 and FID is 0. As convex combination weight $\alpha$ of the reconstructed images increase, there is a gradual deterioration in the image quality, which decreases the accuracy and conversely increases FID. This further confirms that the error in decoder has significant influence on the two metrics (FID and accuracy). Though the purpose here is to demonstrate the generation ability of ReGene framework, it should be noted that with an improved decoder (implying lower FID for Org vs Recon), ReGene has the potential to generate images with even lower FID scores (for Org vs Gen).

\subsection{ReGene: Advantages and Disadvantages}
Design – In ReGene two probabilistic models - classifier $p(y|x)$ and decoder $p(x|z)$ are defined and trained separately. Unlike in VAEs and C-VAEs, there is no need for probabilistic modelling of $p(z|x)$ and $p(z|x, y)$ respectively. Also, instead of randomly sampling the latent space as in VAEs/C-VAEs, ReGeNe framework uses convex combinations to generate new samples (Lemma 1).

Advantages – (i) Guarantee for same class generations (from Lemma \ref{lemma1} and Theorem \ref{theorem3}); (ii) No prior required to model latent space distribution; (ii) No mode drop issue (as selection samples for a given class are deterministic in nature), and (iv) Stable and straight-forward training procedure  (no adversarial training required). Disadvantages – (i) Cannot directly evaluate $p(x)$ for the data distribution; (ii) Trade-off between reconstruction and classification accuracy, and (iii) Quality of image generations is dependent on the reconstruction ability of decoder.
\vspace{-0.1cm}
\section{Conclusion and Future Work}
The answer to the question: `Can classification latent space representations be reused for the downstream task of generation?' is \textit{Yes} - through the proposal of a novel ReGene framework using the convex combination of classification latent space representations.
We showed quantitative and qualitative results on standard datasets and demonstrated comparable performance with other existing state-of-the-art methods. While our experiments utilized simple network structures to prove the possibility of this alternative approach, there is an encouragement for the community to design more sophisticated architectures 
for better reconstruction and therefore, generation. Also, since Lemma \ref{lemma1} is generalizable, its application in other domains foresees exciting potential.

\bibliographystyle{splncs04}
\bibliography{egbib}

\begin{thebibliography}{10}
\providecommand{\url}[1]{\texttt{#1}}
\providecommand{\urlprefix}{URL }
\providecommand{\doi}[1]{https://doi.org/#1}

\bibitem{danbooru2018}
Anonymous, community, D., Branwen, G., Gokaslan, A.: Danbooru2018: A
  large-scale crowdsourced and tagged anime illustration dataset.
  \url{https://www.gwern.net/Danbooru2018} (January 2019),
  \url{https://www.gwern.net/Danbooru2018}

\bibitem{arjovsky2017wasserstein}
Arjovsky, M., Chintala, S., Bottou, L.: {W}asserstein generative adversarial
  networks. In: Precup, D., Teh, Y.W. (eds.) Proceedings of the 34th
  International Conference on Machine Learning. Proceedings of Machine Learning
  Research, vol.~70, pp. 214--223. PMLR, International Convention Centre,
  Sydney, Australia (06--11 Aug 2017),
  \url{http://proceedings.mlr.press/v70/arjovsky17a.html}

\bibitem{Bengio_BetterMixing_2013}
Bengio, Y., Mesnil, G., Dauphin, Y., Rifai, S.: Better mixing via deep
  representations. In: Proceedings of the 30th International Conference on
  Machine Learning, {ICML} 2013, Atlanta, GA, USA, 16-21 June 2013. pp.
  552--560 (2013), \url{http://jmlr.org/proceedings/papers/v28/bengio13.html}

\bibitem{Goodfellow_Understanding_Interpol_ICLR2019}
Berthelot, D., Raffel, C., Roy, A., Goodfellow, I.J.: Understanding and
  improving interpolation in autoencoders via an adversarial regularizer. CoRR
  \textbf{abs/1807.07543} (2018), \url{http://arxiv.org/abs/1807.07543}

\bibitem{BerthelotSM2017began}
Berthelot, D., Schumm, T., Metz, L.: {BEGAN:} boundary equilibrium generative
  adversarial networks. CoRR  \textbf{abs/1703.10717} (2017),
  \url{http://arxiv.org/abs/1703.10717}

\bibitem{BrockDS19biggan}
Brock, A., Donahue, J., Simonyan, K.: Large scale {GAN} training for high
  fidelity natural image synthesis. In: 7th International Conference on
  Learning Representations, {ICLR} 2019, New Orleans, LA, USA, May 6-9, 2019
  (2019), \url{https://openreview.net/forum?id=B1xsqj09Fm}

\bibitem{dai2018diagnosing}
Dai, B., Wipf, D.: Diagnosing and enhancing {VAE} models. In: International
  Conference on Learning Representations (2019),
  \url{https://openreview.net/forum?id=B1e0X3C9tQ}

\bibitem{DinhSB2017realnvp}
Dinh, L., Sohl{-}Dickstein, J., Bengio, S.: Density estimation using real
  {NVP}. In: 5th International Conference on Learning Representations, {ICLR}
  2017, Toulon, France, April 24-26, 2017, Conference Track Proceedings (2017),
  \url{https://openreview.net/forum?id=HkpbnH9lx}

\bibitem{Goodfellow-et-al-2016}
Goodfellow, I., Bengio, Y., Courville, A.: Deep Learning. MIT Press (2016),
  \url{http://www.deeplearningbook.org}

\bibitem{GAN_Original_2014}
Goodfellow, I., Pouget-Abadie, J., Mirza, M., Xu, B., Warde-Farley, D., Ozair,
  S., Courville, A., Bengio, Y.: Generative adversarial nets. In: Ghahramani,
  Z., Welling, M., Cortes, C., Lawrence, N.D., Weinberger, K.Q. (eds.) Advances
  in Neural Information Processing Systems 27, pp. 2672--2680. Curran
  Associates, Inc. (2014),
  \url{http://papers.nips.cc/paper/5423-generative-adversarial-nets.pdf}

\bibitem{gulrajani2017improved}
Gulrajani, I., Ahmed, F., Arjovsky, M., Dumoulin, V., Courville, A.C.: Improved
  training of wasserstein gans. In: Advances in neural information processing
  systems. pp. 5767--5777 (2017)

\bibitem{he2016deep}
He, K., Zhang, X., Ren, S., Sun, J.: Deep residual learning for image
  recognition. In: Proceedings of the IEEE conference on computer vision and
  pattern recognition. pp. 770--778 (2016)

\bibitem{heusel2017gans}
Heusel, M., Ramsauer, H., Unterthiner, T., Nessler, B., Hochreiter, S.: Gans
  trained by a two time-scale update rule converge to a local nash equilibrium.
  In: Advances in Neural Information Processing Systems. pp. 6626--6637 (2017)

\bibitem{huang2017densely}
Huang, G., Liu, Z., Van Der~Maaten, L., Weinberger, K.Q.: Densely connected
  convolutional networks. In: Proceedings of the IEEE conference on computer
  vision and pattern recognition. pp. 4700--4708 (2017)

\bibitem{isola2017image}
Isola, P., Zhu, J.Y., Zhou, T., Efros, A.A.: Image-to-image translation with
  conditional adversarial networks. In: Proceedings of the IEEE conference on
  computer vision and pattern recognition. pp. 1125--1134 (2017)

\bibitem{IvanovFV2019variational}
Ivanov, O., Figurnov, M., Vetrov, D.P.: Variational autoencoder with arbitrary
  conditioning. In: 7th International Conference on Learning Representations,
  {ICLR} 2019, New Orleans, LA, USA, May 6-9, 2019 (2019),
  \url{https://openreview.net/forum?id=SyxtJh0qYm}

\bibitem{perceptual_loss}
Johnson, J., Alahi, A., Li, F.F.: Perceptual losses for real-time style
  transfer and super-resolution  (03 2016)

\bibitem{Prog_GAN_2018}
Karras, T., Aila, T., Laine, S., Lehtinen, J.: Progressive growing of gans for
  improved quality, stability, and variation. CoRR  \textbf{abs/1710.10196}
  (2017), \url{http://arxiv.org/abs/1710.10196}

\bibitem{KingmaW2013autoencoding}
Kingma, D.P., Welling, M.: Auto-encoding variational bayes. In: 2nd
  International Conference on Learning Representations, {ICLR} 2014, Banff, AB,
  Canada, April 14-16, 2014, Conference Track Proceedings (2014),
  \url{http://arxiv.org/abs/1312.6114}

\bibitem{kingma2018glow}
Kingma, D.P., Dhariwal, P.: Glow: Generative flow with invertible 1x1
  convolutions. In: Advances in Neural Information Processing Systems. pp.
  10215--10224 (2018)

\bibitem{Krizhevsky09learningmultiple}
Krizhevsky, A.: Learning multiple layers of features from tiny images. Tech.
  rep. (2009)

\bibitem{KrizhevskySH2017alexnet}
Krizhevsky, A., Sutskever, I., Hinton, G.E.: Imagenet classification with deep
  convolutional neural networks. Commun. {ACM}  \textbf{60}(6),  84--90 (2017).
  \doi{10.1145/3065386}, \url{http://doi.acm.org/10.1145/3065386}

\bibitem{lecun2010mnist}
LeCun, Y., Cortes, C., Burges, C.: Mnist handwritten digit database. ATT Labs
  [Online]. Available: http://yann. lecun. com/exdb/mnist  \textbf{2} (2010)

\bibitem{conf/iccv/LiuLWT15}
Liu, Z., Luo, P., Wang, X., Tang, X.: Deep learning face attributes in the
  wild. In: ICCV. pp. 3730--3738. IEEE Computer Society (2015),
  \url{http://dblp.uni-trier.de/db/conf/iccv/iccv2015.html#LiuLWT15}

\bibitem{MakhzaniF2014ksparse}
Makhzani, A., Frey, B.J.: k-sparse autoencoders. In: 2nd International
  Conference on Learning Representations, {ICLR} 2014, Banff, AB, Canada, April
  14-16, 2014, Conference Track Proceedings (2014),
  \url{http://arxiv.org/abs/1312.5663}

\bibitem{mirza2014conditional}
Mirza, M., Osindero, S.: Conditional generative adversarial nets. arXiv
  preprint arXiv:1411.1784  (2014)

\bibitem{ng2011sparse}
Ng, A., et~al.: Sparse autoencoder. CS294A Lecture notes  \textbf{72}(2011),
  1--19 (2011)

\bibitem{Vladik2018whymixture}
Pownuk, A., Kreinovich, V.: Combining Interval, Probabilistic, and Other Types
  of Uncertainty in Engineering Applications, Studies in Computational
  Intelligence, vol.~773. Springer (2018). \doi{10.1007/978-3-319-91026-0},
  \url{https://doi.org/10.1007/978-3-319-91026-0}

\bibitem{DCGAN_Radford_ICLR2016}
Radford, A., Metz, L., Chintala, S.: Unsupervised representation learning with
  deep convolutional generative adversarial networks. In: {ICLR} (2016)

\bibitem{ali2019vqvae2}
Razavi, A., van~den Oord, A., Vinyals, O.: Generating diverse high-fidelity
  images with {VQ-VAE-2}. CoRR  \textbf{abs/1906.00446} (2019),
  \url{http://arxiv.org/abs/1906.00446}

\bibitem{Sainburg_Adv_Interpol_AE_2018}
Sainburg, T., Thielk, M., Theilman, B., Migliori, B., Gentner, T.: Generative
  adversarial interpolative autoencoding: adversarial training on latent space
  interpolations encourage convex latent distributions. CoRR
  \textbf{abs/1807.06650} (2018), \url{http://arxiv.org/abs/1807.06650}

\bibitem{SalimansGZCRCC2016improved}
Salimans, T., Goodfellow, I.J., Zaremba, W., Cheung, V., Radford, A., Chen, X.:
  Improved techniques for training gans. In: Advances in Neural Information
  Processing Systems 29: Annual Conference on Neural Information Processing
  Systems 2016, December 5-10, 2016, Barcelona, Spain. pp. 2226--2234 (2016),
  \url{http://papers.nips.cc/paper/6125-improved-techniques-for-training-gans}

\bibitem{ShmelkovSA2018how}
Shmelkov, K., Schmid, C., Alahari, K.: How good is my gan? In: Computer Vision
  - {ECCV} 2018 - 15th European Conference, Munich, Germany, September 8-14,
  2018, Proceedings, Part {II}. pp. 218--234 (2018).
  \doi{10.1007/978-3-030-01216-8\_14},
  \url{https://doi.org/10.1007/978-3-030-01216-8\_14}

\bibitem{sohn2015learning}
Sohn, K., Lee, H., Yan, X.: Learning structured output representation using
  deep conditional generative models. In: Advances in neural information
  processing systems. pp. 3483--3491 (2015)

\bibitem{szegedy2016rethinking}
Szegedy, C., Vanhoucke, V., Ioffe, S., Shlens, J., Wojna, Z.: Rethinking the
  inception architecture for computer vision. In: Proceedings of the IEEE
  conference on computer vision and pattern recognition. pp. 2818--2826 (2016)

\bibitem{tolstikhin2018wasserstein}
Tolstikhin, I., Bousquet, O., Gelly, S., Schoelkopf, B.: Wasserstein
  auto-encoders. In: International Conference on Learning Representations
  (2018), \url{https://openreview.net/forum?id=HkL7n1-0b}

\bibitem{WangBSS2004ssim}
Wang, Z., Bovik, A.C., Sheikh, H.R., Simoncelli, E.P.:

\bibitem{DBLP:journals/corr/abs-1708-07747}
Xiao, H., Rasul, K., Vollgraf, R.: Fashion-mnist: a novel image dataset for
  benchmarking machine learning algorithms. CoRR  \textbf{abs/1708.07747}
  (2017), \url{http://arxiv.org/abs/1708.07747}

\bibitem{ZhangZLBP2019perceptual}
Zhang, Z., Zhang, R., Li, Z., Bengio, Y., Paull, L.: Perceptual generative
  autoencoders. In: Deep Generative Models for Highly Structured Data, {ICLR}
  2019 Workshop, New Orleans, Louisiana, United States, May 6, 2019 (2019),
  \url{https://openreview.net/forum?id=rkxkr8UKuN}

\end{thebibliography}
\end{document}